\documentclass[10pt,twocolumn,letterpaper]{article}

\usepackage{cvpr}              
\usepackage{multirow}
\usepackage{graphicx}
\usepackage{tcolorbox}
\usepackage{tabularx}
\usepackage[table]{xcolor}
\usepackage{longtable}
\usepackage{makecell}
\usepackage{tcolorbox}
\tcbuselibrary{breakable}
\usepackage{colortbl}
\definecolor{tablegray}{gray}{0.9}
\definecolor{cvprblue}{rgb}{0.21,0.49,0.74}

\usepackage[pagebackref,breaklinks,colorlinks,allcolors=cvprblue]{hyperref}

\title{
Refer-Agent: A Collaborative Multi-Agent System with Reasoning and Reflection for Referring Video Object Segmentation
}

\author{
\textbf{Haichao Jiang\textsuperscript{1,\thanks{Equal contribution.}} \ \
Tianming Liang\textsuperscript{1,\footnotemark[1]}  \ \ 
Wei-Shi Zheng\textsuperscript{1} \ \ 
Jian-Fang Hu\textsuperscript{1}\thanks{Corresponding author.}}\\
\textsuperscript{1}Sun Yat-sen University \ \ \\
}

\begin{document}
\maketitle
\begin{abstract}
Referring Video Object Segmentation (RVOS) aims to segment objects in videos based on textual queries. 
Current methods mainly rely on large-scale supervised fine-tuning (SFT) of Multi-modal Large Language Models (MLLMs). However, this paradigm suffers from heavy data dependence and limited scalability against the rapid evolution of MLLMs. 
Although recent zero-shot approaches offer a flexible alternative, their performance remains significantly behind SFT-based methods, due to the straightforward workflow designs. 
To address these limitations, we propose \textbf{Refer-Agent}, a collaborative multi-agent system with alternating reasoning-reflection mechanisms. This system decomposes RVOS into step-by-step reasoning process. During reasoning, we introduce a Coarse-to-Fine frame selection strategy to ensure the frame diversity and textual relevance, along with a Dynamic Focus Layout that adaptively adjusts the agent's visual focus.
Furthermore, we propose a Chain-of-Reflection mechanism, which employs a Questioner-Responder pair to generate a self-reflection chain, enabling the system to verify intermediate results and generates feedback for next-round reasoning refinement. 
Extensive experiments on five challenging benchmarks demonstrate that Refer-Agent significantly outperforms state-of-the-art methods, including both SFT-based models and zero-shot approaches.
Moreover, Refer-Agent is flexible and enables fast integration of new MLLMs without any additional fine-tuning costs. Code will be released at \url{https://github.com/iSEE-Laboratory/Refer-Agent}.
\end{abstract}

\section{Introduction}
\label{sec:intro}

\begin{figure}[h]
\vspace{-1.5em}
    \centering
    \includegraphics[width=0.45\textwidth]{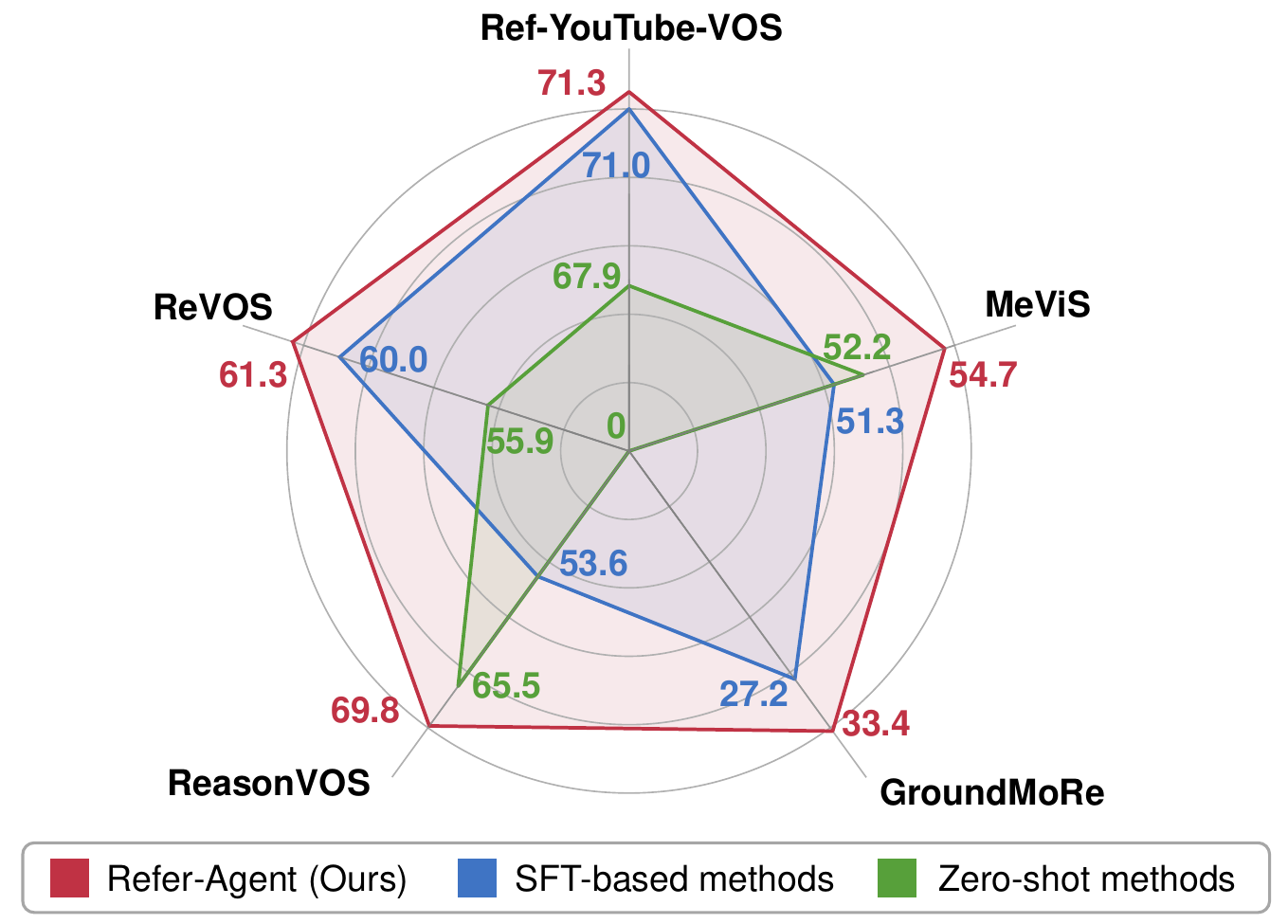}
    \vspace{-0.7em}
    \caption{\textbf{Comparison with SOTAs}. Without any fine-tuning, our Refer-Agent achieves the best performances across five RVOS datasets and outperforms all previous state-of-the-art methods, including both zero-shot approaches and SFT-based models.}
    \label{fig:radar}
    \vspace{-1.5em}
\end{figure}
Referring Video Object Segmentation (RVOS)~\cite{botach2022end, wu2022referformer} has emerged as a popular research topic in recent years. This task aims to segment the target objects in a video based on a textual query. However, user queries are typically diverse and unpredictable, ranging from straightforward descriptions (e.g., ``the man in the blue shirt'') to complex instructions (e.g, ``Can you identify the individual that appears to be excluded from the group?''). In order to understand such complicated user intention and associate it with the visual objects accurately, RVOS models must possess strong vision-language understanding and reasoning capabilities.

To address this challenge, recent efforts have begun integrating Multimodal Large Language Models (MLLMs) into RVOS pipelines. These approaches typically customize a set of learnable semantic embeddings to bridge MLLMs and segmentation models (e.g., SAM~\cite{kirillov2023segment} and SAM2~\cite{ravi2024sam}), thereby enabling end-to-end joint training. However, this paradigm is significantly inefficient in practice, because both MLLMs and foundational segmentation models are inherently data-hungry. Researchers have to collect extensive data from multiple sources and consume considerable computational resources to conduct large-scale supervised fine-tuning (SFT).
Nowadays, MLLMs are undergoing rapid evolution, with new architectures 
emerging frequently~\cite{bai2025qwen2,liu2023visual,yang2025kwai,hong2025glm}. Repeating the high-cost SFT on each newly released MLLM is inefficient and impractical. 

In this work, we seek a \textbf{training-free and scalable paradigm} that leverages only the zero-shot capabilities of pretrained models to address complex RVOS tasks. This direction has been preliminarily explored in recent studies~\cite{huang2024unleashing,CoTRVS}, which typically combine MLLMs, image grounding models, and segmentation models into a collaborative workflow to accomplish the RVOS task.
However, it is non-trivial to design an effective collaboration mechanism for these disparate models in a training-free manner.
\textbf{1)} Current methods rely solely on a straightforward workflow, with each worker focusing only on its assigned subtask. The lack of self-corrected mechanisms can lead to significant error accumulation throughout the process.
\textbf{2)} MLLMs are inherently prone to hallucinations. They often fabricate non-existent evident and generate confident yet incorrect predictions when handling complex user queries and video scenes. 
Therefore, while promising, the performance of current zero-shot methods is still far from SFT-based methods, which significantly limits their value in practice.

To fill this gap, we propose \textbf{Refer-Agent}, a collaborative multi-agent system with alternating reasoning-reflection mechanisms. 
Specifically, our primary pipeline decomposes the RVOS task into four stages: (i) Frame Selection, (ii) Intent Analysis, (iii) Object Grounding, and (iv) Mask Generation. In contrast to prior single-pass frame sampling~\cite{yan2024visa, bai2024one}, we introduce a \textit{Coarse-to-Fine} frame selection approach to ensure both the frame diversity and visual-text relevance. Furthermore, in the Intent Analysis stage, we propose a \textit{Dynamic Focus Layout} strategy that dynamically allocates blocks of varying sizes to different frames, enabling the agent to maintain focus on the keyframe while retaining essential information from the context frames. 

Although this main pipeline is sufficiently to perform step-by-step object reasoning segmentation, it remains unreliable due to the inherent MLLM hallucinations. To address this limitation, we furhter propose a Chain-of-Reflection (CoR) mechanism, which employs a pair of Questioner and Responder to generate a self-reflection chain, enabling the agent system to verify the intermediate results, identify the errors, and receive instructive feedback for next-round reasoning refinement.  
Together, these components form a multi-round collaborative framework that alternates between reasoning and reflection, ultimately producing robust and accurate segmentation results.

We evaluate our approach on five popular RVOS benchmarks, which involve complex reasoning, world knowledge, and temporal understanding.
Extensive experimental results demonstrate that our Refer-Agent achieves state-of-the-art (SOTA) performance, significantly surpassing not only the previous zero-shot approaches but also the SFT-based models. 
We also conduct in-depth ablation studies to analyze the effectiveness of each component in zero-shot RVOS systems, which we believe can provide novel insights for future development of general agent systems.

Overall, our contributions are summarized as follows:
\begin{itemize}
    \item We propose \textbf{Refer-Agent}, a novel collaborative multi-agent system with alternating reasoning-reflection mechanisms. A \textit{Coarse-to-Fine} frame selection approach and a \textit{Dynamic Focus Layout} strategy are introduced to enhance the reasoning segmentation.
    \item We introduce a \textbf{Chain-of-Reflection (CoR)} mechanism to further enhance the system robustness.
    \item Without fine-tuning, Refer-Agent establishes new SOTA performance on five benchmarks, significantly surpassing both the existing zero-shot and SFT-based methods.
\end{itemize}

\begin{figure*}
    \centering
    \captionsetup{type=figure}
    \includegraphics[width=1.0\linewidth]{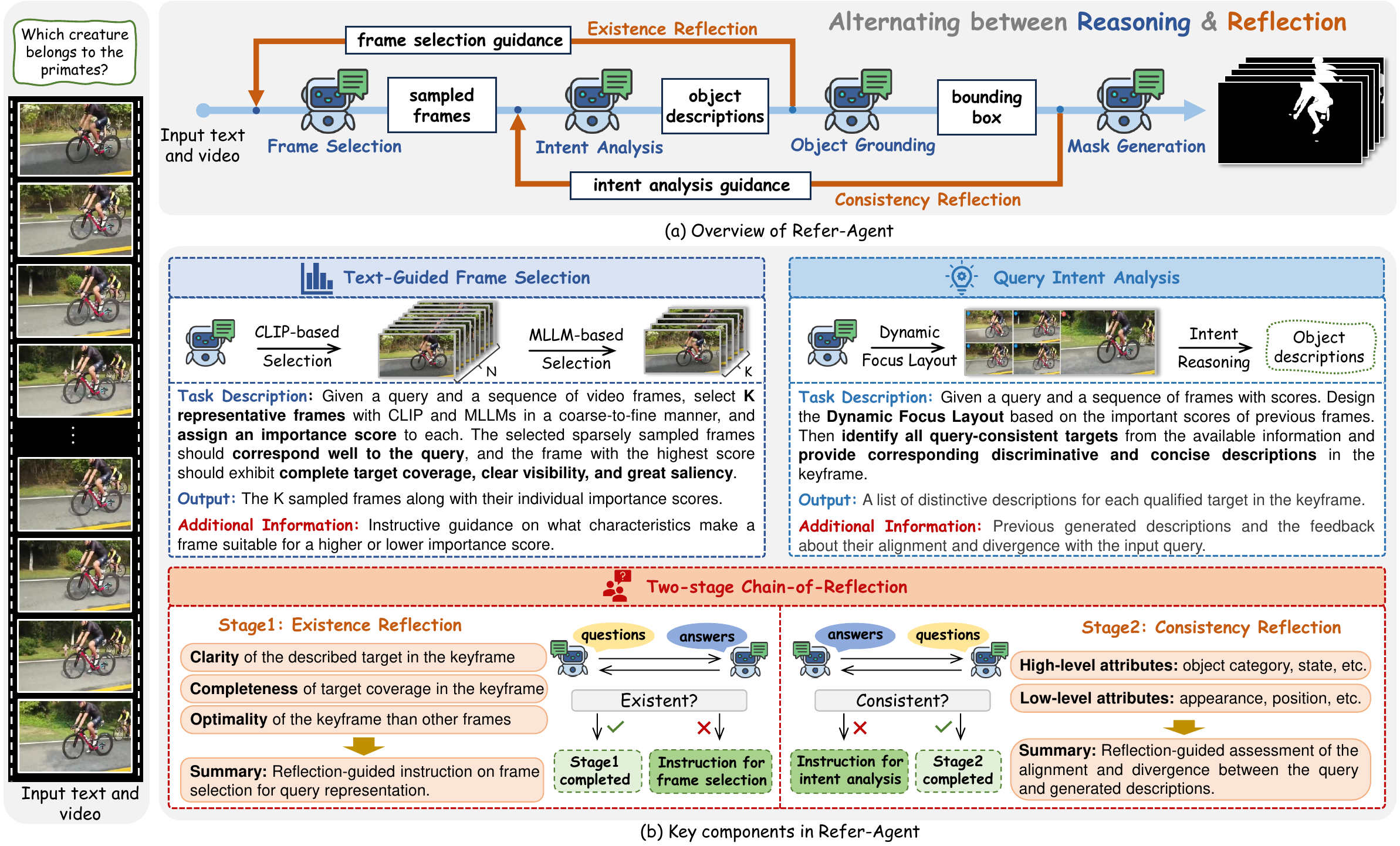}
    \vspace{-1.9em}
    \caption{\textbf{(a)} Overview of the Refer-Agent system. The primary pipeline (\textcolor[HTML]{2F5597}{\textbf{blue}}) performs step-by-step reasoning to achieve object analysis and segmentation, while a Two-stage \textit{Chain-of-Reflection} mechanism (\textcolor[HTML]{C55A11}{\textbf{orange}}) is further integrated to verify and refine the intermediate results. By alternating between reasoning and reflection, our Refer-Agent can produce robust RVOS predictions. 
    \textbf{(b)} Illustration of key components. Specially, the \textit{Chain-of-Reflection} mechanism, comprising Existence Reflection and Consistency Reflection, employs a Questioner-Responder pair to verify intermediate results, identify errors, and provide feedback for next-round reasoning refinement.}
    \label{fig:overall}
    \vspace{-1em}
\end{figure*}

\section{Related Work}
\label{sec:related_work}

\textbf{Referring Video Object Segmentation(RVOS).}
RVOS~\cite{ding2023mevis, seo2020urvos} aims to segment target objects within a video based on given language description. 
Pioneer approaches \cite{botach2022end, wu2022language, han2023html, luo2023soc, miao2023spectrum, yuan2024losh} primarily adopted the DETR~\cite{carion2020end} framework, focusing on identifying objects using simple appearance descriptions.
To enhance the reasoning capability of RVOS models, recent studies~\cite{ yan2024visa, bai2024one, lin2025glus, sa2va, wang2025object,
gong2025devil} seek to integrate multi-modal large language models (MLLMs) with foundational segmentation models with large-scale supervised fine-tuning.
For instance, GLUS~\cite{lin2025glus} constructs unified global and local reasoning within MLLMs.
while RGA3~\cite{wang2025object} unifies video grounding and segmentation in a multi-round conversational framework.
However, this paradigm requires large-scale training data and substantial computational resources for SFT, making it difficult to keep pace with the rapid evolution of MLLMs.
Therefore, recent research has begun to explore training-free RVOS methods. AL-Ref-SAM2~\cite{huang2024unleashing} adopts GPT-4 to select pivot frames and boxes. CoT-RVS~\cite{CoTRVS} employs MLLMs for frame selection and referring image segmentation models for mask generation. While promising, current zero-shot methods are still limited by the straightforward workflow designs and their performance remains significantly behind SFT-based methods. In this work, we propose Refer-Agent to fill this critical gap. By integrating a novel alternating reasoning-reflection mechanism, Refer-Agent overcomes the limitations of current zero-shot approaches, establishing new SOTA performance across five challenging benchmarks without any fine-tuning.

\noindent \textbf{Video Object Segmentation(VOS).}
Starting from an initial object mask in the first frame, VOS aims to track the object and propagate the mask throughout the video~\cite{pont20172017}. The most impressive work in recent years is SAM2~\cite{ravi2024sam}, which introduces a unified prompt-based segmentation paradigm, enabling users to prompt objects with points, box, or mask. SAM2 has sparked considerable follow-up research~\cite{ding2025sam2long, yang2024samurai, zhang2025sec, videnovic2025distractor}. SAM2Long~\cite{ding2025sam2long} uses tree-based memory to enhance object tracking in long video. SeC~\cite{zhang2025sec} additionally employs MLLMs to generate conceptual priors for SAM2. These methods focus on visual-prompting segmentation, while our work targets segmentation by text prompts.

\noindent\textbf{Vision Agent.}
Many recent studies explore to utilize the zero-shot capabilities of agents to address various complex vision tasks~\cite{fan2024videoagent, wang2025videotree, chen2025lvagent, chen2025edival}. EDIVAL-AGENT~\cite{chen2025edival} designs an automated evaluation agent for multi-turn instruction-based image editing. LVAgent~\cite{chen2025lvagent} employs a multi-round dynamic collaboration of MLLM agents for long video understanding. In this work, we aim to explore the agent system for RVOS, as it involves complicated cross-modal reasoning. Our proposed Chain-of-Reflection mechanism can effectively mitigate the MLLM hallucinations and error accumulation during the reasoning process, enabling our Refer-Agent to achieve surprising zero-shot performance, even surpassing those SFT-based methods significantly.

\section{Refer-Agent}
\label{sec:method}

\noindent\textbf{Overview.}
Given a textual query $x_q$ and a video $x_v=\{f_t\}_{t=1}^T \in \mathbb{R}^{T \times H \times W \times 3}$, where $T$ is the frame number and each frame $f_t$ has a size of $H \times W$,
the goal of RVOS is to segment the referred objects throughout the video and output a list of masklets $\mathcal{M}=\{\mathcal{M}_i \in \mathbb{R}^{T \times H \times W}\}_{i=1}^k$, where $k$ is the number of referred objects.
As illustrated in Figure~\ref{fig:overall}, our Refer-Agent integrates a primary reasoning-segmentation pipeline with a Chain-of-Reflection (CoR) mechanism.
The main pipeline performs step-by-step reasoning segmentation, which involves frame selection, intent analysis, object grounding, and mask generation.
In parallel, the CoR mechanism is designed to capture potential hallucinations and iteratively refine intermediate predictions through a two-stage verification process.
This process continues iteratively until: (1) both stages of verification are passed, or (2) the maximum number of reflection is reached.

\subsection{Step-by-step Reasoning Pipeline}
\label{sec:pipeline}
As illustrated in Figure \ref{fig:overall}, the step-by-step reasoning segmentation process in our Refer-Agent involves a chain of agents to sequentially perform frame selection, intent analysis, object grounding, and mask generation.

\vspace{0.5em}\noindent
\textbf{Text-guided Frame Selection.} 
This agent aims to sample a set of representative frames from the raw video for efficient segmentation reasoning, because performing reasoning on dense frames is typically expensive and noisy.
To ensure the diversity and text-relevance of sampled frames, we introduce a text-guided frame selection strategy that selects $K$ representative frames in a coarse-to-fine manner. Firstly, we coarsely select $N$ frames ($N\gg K$) from the entire video by computing the CLIP-based semantic similarity \( S_\text{CLIP} \) between the textual query \( x_q \) and each frame. We then divide the video into \( N \) segments, and collect the frame with the highest score in each segment, resulting in a representative frame set \( X_{\text{sample}} \in \mathbb{R}^{N \times H \times W} \).

Subsequently, we perform the fine frame selection by feeding the sampled frames $X_\text{sample}$ and the textual queries $x_q$ into an MLLM, prompting it to analyze the video content and assign an importance score to each frame, reflecting how well it represents the potential target objects.
Finally, we select top-$K$ frames as output based on the following score:
\vspace{-2mm}
\begin{equation}
    S^t = \alpha \cdot S_\text{CLIP}^t + \beta \cdot S_{\text{MLLM}}^t,
    \vspace{-2mm}
    \label{eq:score}
\end{equation}
where $S_\text{CLIP}^t$ and $S_\text{MLLM}^t$ are the CLIP and MLLM scores of $t$-th frame, respectively. And $\alpha$ and $\beta$ are hyper-parameters. 

It is worth noting that while our coarse-to-fine strategy ensures better frame diversity and visual-text alignment than previous single-pass frame sampling methods~\cite{bai2024one,yan2024visa}, both approaches lack robustness in complex reasoning scenarios. This is because the frame selection currently depends only on the original query.
For example, for query ``who makes the girl cry", the current strategy will prioritize frames for the \textit{crying girl}, rather than the \textit{cause} of this event. To address the limitation, we further propose a multi-round reasoning-enhanced strategy, which progressively refines frame selection by reasoning-reflection (Section~\ref{sec:reflection}). 

\begin{figure}
    \centering
    \captionsetup{type=figure}
    \includegraphics[width=\linewidth]{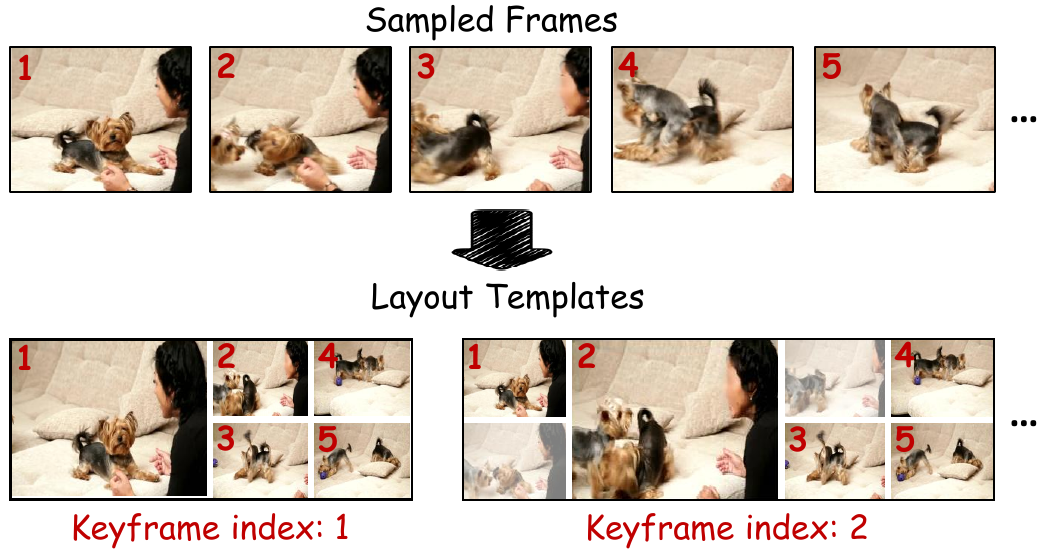}
    \vspace{-1.8em}
    \caption{Illustration of Dynamic Focus Layout. The number within each block denotes the frame index.}
    \label{fig:combined}
    \vspace{-1.5em}
\end{figure}

\vspace{0.5em}\noindent\textbf{Intent Analysis.}
User queries are often obscure. To capture the target accurately, it is necessary to analyze the query intent and confirm the targets before localizing them.
Given the selected frames and the query (e.g., ``the persons who are appreciating music"), this agent is instructed to perform object reasoning and generate a concise yet discriminative expression (e.g., ``the man in blue") for each target. 

Early works~\cite{huang2024unleashing} only perform reasoning on a single keyframe, which is often ineffective due to the lack of temporal context. 
Recent works~\cite{CoTRVS} input all sampled frames equally to the MLLM. However, this is computationally inefficient. Besides, the large number of input frames can disperse the MLLM's attention, thereby causing hallucinations.
To highlight the keyframe while keeping the context information, we propose \textit{Dynamic Focus Layout}, which dynamically allocates a larger, higher-resolution block to the keyframe and compresses the remaining context frames into smaller, low-resolution slots. 
Note that the keyframe is defined as the frame with highest score.
As illustrated in Figure~\ref{fig:combined},  in order to maintain the temporal information, the layout template is determined by the keyframe index. If the keyframe index is even, we sample two additional frames from the original video to maintain a balanced grid structure. This composed image and the textual query are then fed into the MLLM, which is prompted to reason over the video content, identify relevant objects, and generate an expression for each predicted target in the keyframe.

\vspace{0.3em}\noindent\textbf{Object Grounding and Segmentation.}
MLLMs are typically capable of image grounding~\cite{bai2025qwen2,lu2025ovis2}.
For each target, the MLLM takes the keyframe and the generated expression as input, and localize the target with bounding box coordinates. 
These bounding boxes, along with the keyframe, are ultimately used as the initial prompt for a video segmentation model (e.g., SAM2~\cite{ravi2024sam}), which generates precise object masks and propagates them throughout the entire video.

\renewcommand{\arraystretch}{1}
\begin{table*}[t]
	\caption{Comparison with state-of-the-art methods on Ref-YouTube-VOS, MeViS, ReasonVOS and GroundMoRe datasets.}
        \vspace{-1em}
	\centering
    \setlength{\tabcolsep}{6.5pt}
    \resizebox{\textwidth}{!}{
	   \begin{tabular}{l|c|ccc|ccc|ccc|ccc}
			\hline
			\multirow{2}{*}{Method} & \multirow{2}{*}{MLLM} & \multicolumn{3}{c|}{Ref-YouTube-VOS} & \multicolumn{3}{c|}{MeViS} & \multicolumn{3}{c|}{ReasonVOS} &\multicolumn{3}{c}{GroundMoRe} \\
			& & $\mathcal{J}$\&$\mathcal{F}$ & $\mathcal{J}$ & $\mathcal{F}$ & $\mathcal{J}$\&$\mathcal{F}$ & $\mathcal{J}$ & $\mathcal{F}$ & $\mathcal{J}$\&$\mathcal{F}$ & $\mathcal{J}$ & $\mathcal{F}$ & $\mathcal{J}$\&$\mathcal{F}$ & $\mathcal{J}$ & $\mathcal{F}$\\
			\hline
            \rowcolor{gray!7.5}
            \multicolumn{14}{l}{\emph{SFT-based Methods}} \\
			\hline
            VISA~\cite{yan2024visa} & Chat-UniVi-13B & 63.0 & 61.4 & 64.7 & 44.5 & 41.8 & 47.1 & -- & -- & -- & 5.3 & 4.7 & 5.9 \\
                VideoLISA~\cite{bai2024one} & LLaVA-3.8B & 63.7 & 61.7 & 65.7 & 44.4 & 41.3 & 47.6 & 47.5 & 45.1 & 49.9 & -- & -- & -- \\
			    GLUS~\cite{lin2025glus} & LISA-7B & 67.3 & 65.5 & 69.0 & 51.3 & 48.5 & 54.2 & 49.9 & 47.5 & 52.4 & -- & -- & -- \\
                Sa2VA~\cite{sa2va} & InternVL2-8B & 70.7 & -- & -- & 46.9 & -- & -- & -- & -- & -- & -- & -- & -- \\
                MoRa~\cite{deng2025motion} & LISA-7B & -- & -- & -- & -- & -- & -- & -- & -- & -- & 27.2 & 27.4 & 26.9 \\
                RGA3~\cite{wang2025object} & Qwen2.5-VL-7B & 68.5 & 66.8 & 70.1 & 50.1 & 47.4 & 52.8 & 53.6 & 51.3 & 56.0 & -- & -- & -- \\
                VRS-HQ~\cite{gong2025devil} & Chat-UniVi-13B & 71.0 & \textbf{69.0} & 73.1 & 50.9 & 48.0 & 53.7 & -- & -- & -- & -- & -- & -- \\
			\hline
            \rowcolor{gray!7.5}
            \multicolumn{14}{l}{\emph{Zero-shot Methods}} \\
			\hline
                AL-Ref-SAM2~\cite{huang2024unleashing} & GPT4 & 67.9 & 65.9 & 69.9 & 42.8 & 39.5 & 46.2  & -- & -- & -- & -- & -- & -- \\
			    CoT-RVS~\cite{CoTRVS} & Gemma3-12B & - & - & - & 44.2 & 40.3 & 48.1 & 50.7 & 47.5 &54.0 & -- & -- & -- \\
                CoT-RVS~\cite{CoTRVS} & GPT-4o & - & - & - & 52.2 & 48.7 & 55.7 & 65.5 & 62.4 & 68.7 & -- & -- & -- \\
			\hline
            \rowcolor{blue!7.5}
                \textbf{Refer-Agent} (ours) & Ovis2.5-9B & \textbf{71.3} & \textbf{69.0} & \textbf{73.6} & \textbf{54.7} & \textbf{51.6} & \textbf{57.7} & \textbf{69.8} & \textbf{67.0} & \textbf{72.7} & \textbf{33.4} & \textbf{32.0} & \textbf{35.0} \\
			\hline
	   \end{tabular}
    }
	\label{tab:other_datasets}
    \vspace{-0.5em}
\end{table*}

\renewcommand{\arraystretch}{1}
\begin{table*}[!t]
        \caption{Comparison with state-of-the-art methods on ReVOS dataset.}
                \vspace{-1em}
    \centering
    \setlength{\tabcolsep}{9pt}
    \resizebox{\textwidth}{!}{
        \begin{tabular}{l|c|ccc|ccc|ccc|c}
            \hline
            \multirow{2}{*}{Method} & \multirow{2}{*}{MLLM} & \multicolumn{3}{c|}{Referring} & \multicolumn{3}{c|}{Reasoning} & \multicolumn{3}{c|}{Overall} & \multirow{2}{*}{$\mathcal{R}$} \\
            & & $\mathcal{J}$\&$\mathcal{F}$ & $\mathcal{J}$ & $\mathcal{F}$ & $\mathcal{J}$\&$\mathcal{F}$ & $\mathcal{J}$ & $\mathcal{F}$ & $\mathcal{J}$\&$\mathcal{F}$ & $\mathcal{J}$ & $\mathcal{F}$ &  \\
            \hline 
            \rowcolor{gray!7.5}
            \multicolumn{12}{l}{\emph{SFT-based Methods}} \\
            \hline
            VISA~\cite{yan2024visa} & Chat-UniVi-13B & 57.4 & 55.6 & 59.1 & 44.3 & 42.0 & 46.7 & 50.9 & 48.8 & 52.9 & 14.5 \\
            HyperSeg~\cite{wei2025hyperseg} & Mipha-3B & 58.5 & 56.0 & 60.9 & 53.0 & 50.2 & 55.8 & 55.7 & 53.1 & 58.4 & -- \\
            InstructSeg~\cite{wei2025instructseg} & Mipha-3B & 57.0 & 54.8 & 59.2 & 51.9 & 49.2 & 54.7 & 54.5 & 52.0 & 56.9 & -- \\
            GLUS~\cite{lin2025glus} & LISA-7B & 58.3 & 56.0 & 60.7 & 51.4 & 48.8 & 53.9 & 54.9 & 52.4 & 57.3 & 17.9 \\
            ViLLa~\cite{zheng2025villa} & InternVideo2-6B & -- & -- & -- & -- & -- & -- & 57.0 & 54.9 & 59.1 & -- \\
            RGA3~\cite{wang2025object} & Qwen2.5-VL-7B & 60.5 & 58.7 & 62.3 & 55.4 & 53.1 & 57.7 & 58.0 & 55.9 & 60.0 & \textbf{28.6} \\
            VRS-HQ~\cite{gong2025devil} & Chat-UniVi-13B & 63.3 & 61.1 & 65.5 & 56.8 & 54.1 & 59.4 & 60.0 & 57.6 & 62.5 & 18.9 \\
            \hline
            \rowcolor{gray!7.5}
            \multicolumn{12}{l}{\emph{Zero-shot Methods}} \\
            \hline
            CoT-RVS~\cite{CoTRVS} & Gemma3-12B & -- & -- & -- & -- & -- & -- & 47.1 & 43.4 & 50.9 & --  \\
            CoT-RVS~\cite{CoTRVS} & GPT-4o & -- & -- & -- & -- & -- & -- & 55.9 & 52.8 & 59.0 & -- \\
            \hline
            \rowcolor{blue!7.5}
            \textbf{Refer-Agent} (ours) & Ovis2.5-9B & \textbf{65.0} & \textbf{62.3} & \textbf{67.7} & \textbf{57.7} & \textbf{54.3} & \textbf{61.0} & \textbf{61.3} & \textbf{58.3} & \textbf{64.3} & \textbf{28.6} \\
            \hline
        \end{tabular}
    }
    \vspace{-2mm}
    \label{tab:ReVOS_results}
\end{table*}

\subsection{Chain-of-Reflection Mechanism}
\label{sec:reflection}
While the main pipeline enables the MLLM to perform step-by-step reasoning and produce corresponding object predictions, the results remain unreliable in complex scenarios. 
This is primarily due to the hallucination issues inherent in MLLMs, which leads them to deviate from the provided video content and fabricate non-existent evidence, resulting in confident yet incorrect predictions.
Due to the lack of a self-corrected mechanism, the initial hallucinations, such as suboptimal frame selection or inaccurate reasoning, can be propagated and amplified throughout the pipeline, significantly affecting the final predictions. 
To address this limitation, we propose CoR. This mechanism comprises two stages: \textit{Existence Reflection} and \textit{Consistency Reflection}. In each stage, a Questioner-Responder pair generates a series of question-answering chains to verify the intermediate correctness, identify the errors, and provide instructive feedback for next-round reasoning refinement.

\vspace{0.3em}\noindent\textbf{Existence Reflection.}
The purpose of this stage is to review the quality of the selected frames, and provide guidelines for enhancing frame selection in next round. 
This is particularly critical in the initial round, where the frames are selected based solely on the original query. In these frames, the ground-truth targets may appear unclear or even absent. Such incomplete visual information makes MLLM reasoning more prone to hallucinations. Therefore, in this stage, the Questioner is instructed to generate the three aspects of questions.
1) \textbf{Visibility:} Are the targets are clearly visible in the keyframe?
2) \textbf{Completeness:} Does the keyframe cover all the referring targets?
3) \textbf{Optimality:} Is there a better choice in the context frames than the current keyframe?
More details about the question templates are provided in the Supplementary. Based on these questions, the Responder re-evaluates the selected frames against the user query and generated expressions. During the question-answering process, a reflection chain is gradually generated. If the verification fails, this reflection chain is fed back into the reasoning pipeline. The frame-selection agent is then prompted to perform a new-round of frame selection, explicitly guided by the feedback. For instance, if the reflection states, ``\textit{The target `white sedan' is occluded in the keyframe}", the agent will search for a new keyframe where the target is more clearly visible.
This self-reflection step ensures that the subsequent reasoning process is built upon high-quality, relevant visual evidence, thereby mitigating hallucinations that arise from incomplete context.

\noindent\textbf{Consistency Reflection.}
This stage verifies the consistency between predicted objects and original query, thereby mitigating hallucinations that may arise during long-horizon reasoning. 
Specifically, the Questioner is prompted to decompose raw query into a set of attributes. Based on these attributes, it then generates a series of multi-choice questions encompassing both the \textit{high-level concepts} (e.g., object category and state) and \textit{low-level details} (e.g., appearance, shape, and spatial location). By reviewing keyframes with predicted objects highlighting via bounding boxes, the Responder answers each question and provide an explanation to justify its choice. After verifying the consistency between answers and ground-truth, a report is summarized and fed back into the reasoning pipeline, unless all answers are correct. For instance, if the referring target is specified as \textit{blue}, but the predicted object is identified as \textit{red}, the intent-reasoning agent can recognize that the color attribute was overlooked in prior reasoning, and revise it in the subsequent round.
This reflection step ensures that all predicted targets are fully faithful to the user provided query, thereby enhancing the reliability of the entire framework.

\section{Experiments}
\label{experiments}
\subsection{Experimental Setup}
\textbf{Datasets.}
Our method is tested on five major language-guided video object segmentation datasets: ReVOS~\cite{yan2024visa}, Ref-YouTube-VOS~\cite{seo2020urvos}, MeViS~\cite{ding2023mevis}, ReasonVOS~\cite{bai2024one} and GroundMoRe~\cite{deng2025motion}. These datasets pose diverse challenges:  ReVOS and ReasonVOS advance Reasoning VOS with queries requiring strong reasoning capabilities and world knowledge; Ref-YouTube-VOS is widely adopted due to its large scale; MeViS is known for its complex scenarios involving multiple similar objects and cross-modal motion understanding; GroundMoRe proposes a new video task called MotionGrounded Video Reasoning for comprehensive motion understanding. Notably, in ReVOS and MeViS, a single query may refer to multiple distinct target objects.

\noindent\textbf{Evaluation Metrics.}
We adopt the metrics used in previous works, including region similarity $\mathcal{J}$ (average IoU), contour accuracy $\mathcal{F}$ (mean boundary similarity), and average $\mathcal{J} \& \mathcal{F}$. 

\noindent\textbf{Implementation Details.}
We adopt Ovis2.5-9B~\cite{lu2025ovis2} as our baseline MLLM, and employ SAM2 \cite{ravi2024sam} as our segmentation agent. For each video, $N=10, K=5$ frames are used in the frame selection stage. The parameters $\alpha$ and $\beta$ are set as $0.3$, and $0.7$, respectively.
The maximum number of reflection \texttt{MAX\_TURN} is set to 4 by default. 
In the Consistency Reflection stage, for multi-object cases, the validation is regarded failed if more than $30\%$ of predicted targets are inconsistent to the ground-truth. 
Our approach is training-free and computationally-friendly, requiring only 8 NVIDIA 3090 GPUs for all experiments.

\renewcommand{\arraystretch}{1}
\begin{table}[t]
    \caption{Evaluation of the two-stage Chain-of-Reflection: ``Stage1''  for Existence and ``Stage2''  for Consistency Reflection.
    }
            \vspace{-3mm}
    \setlength{\tabcolsep}{12pt}
    \centering
    \resizebox{0.45\textwidth}{!}{
        \begin{tabular}{l|ccc}
            \hline
            Method & $\mathcal{J}$\&$\mathcal{F}$ & $\mathcal{J}$ & $\mathcal{F}$ \\
             \hline
             \rowcolor{blue!7.5}
             \textbf{Refer-Agent} & \textbf{69.8} & \textbf{67.0} & \textbf{72.7}\\
             \hline
             w/o Stage1 & 66.6 & 63.7 & 69.5 \\
             w/o Stage2 & 65.2 & 62.4 & 68.1 \\
             w/o Stage1 \& Stage2 & 64.5 & 61.8 & 67.2 \\
            \hline
        \end{tabular}
        }
    \label{table:reflection_ablation}
\end{table}

\renewcommand{\arraystretch}{1}
\begin{table}[t]
    \vspace{-2mm}
    \caption{Ablation study of visual input strategy for intent analysis.}
    \vspace{-3mm}
    \setlength{\tabcolsep}{11pt}
    \centering
    \resizebox{0.45\textwidth}{!}{
        \begin{tabular}{l|ccc}
             \hline
             Strategy & $\mathcal{J}$\&$\mathcal{F}$ & $\mathcal{J}$ & $\mathcal{F}$ \\
             \hline
             Single Keyframe & 67 & 63.8 & 70.2 \\
             Uniform Integration & 67.7 & 64.8 & 70.7 \\
             \hline
             \rowcolor{blue!7.5}
             \textbf{Ours} & \textbf{69.8} & \textbf{67.0} & \textbf{72.7} \\
             \hline
        \end{tabular}
    }
    \label{tab:input_strategy}
    \vspace{-1em}
\end{table}

\subsection{Main Results}
As shown in Tables \ref{tab:other_datasets} and \ref{tab:ReVOS_results}, across all five benchmarks, Refer-Agent consistently outperforms all SOTA competitors, including MLLMs with supervised fine-tuning and the zero-shot methods combining top-tier closed-source models such as GPT-4o.
Without any fine-tuning, Refer-Agent outperforms SFT-based methods by $16.2\%$ and $1.3\%$ on the challenging reasoning datasets ReasonVOS and ReVOS, respectively. Furthermore, in contrast to fine-tuning-based approaches, Refer-Agent enables immediate integration of newly released MLLMs, without any extra training costs.

Compared to previous zero-shot methods that even integrate powerful closed-source MLLMs, our Refer-Agent achieves significant improvements across all the benchmarks with only a 9B open-source model. Specifically, Refer-Agent outperforms CoT-RVS with GPT-4o by $5.4\%$ and $4.3\%$ $\mathcal{J} \& \mathcal{F}$ on ReVOS and ReasonVOS, respectively. These results highlight the effectiveness of our alternating reasoning-reflection mechanisms in mitigating MLLM hallucination when handling complex queries.

On the MeViS and GroundMoRe datasets that demand strong motion and temporal understanding, we can observe that our Refer-Agent still significantly outperform all SOTA methods by $2.5\%$ and $6.2\%$ $\mathcal{J} \& \mathcal{F}$ respectively, which highlights the effectiveness of our approach in understanding complex motion patterns and temporal dependencies.

\renewcommand{\arraystretch}{1}
\begin{table}[t]
    \vspace{-2mm}
    \caption{Influence of agent merging and \texttt{MAX\_TURN}. ``Time'' denotes the processing time (seconds) per sample (excluding video load), estimated on an NVIDIA RTX 3090.
    }
    \vspace{-3mm}
    \setlength{\tabcolsep}{6.5pt}
    \centering
    \resizebox{0.48\textwidth}{!}{
        \begin{tabular}{c|c|ccc|c}
             \toprule
             Method & \texttt{MAX\_TURN} & $\mathcal{J}$\&$\mathcal{F}$ & $\mathcal{J}$ & $\mathcal{F}$ & Time\\
             \hline
             variant1 & 4 & 67.1 & 64.0 & 70.1 & 199 \\
             variant2 & 4 & 65.3 & 62.3 & 68.2 & 182 \\
             variant3 & 4 & 62.5 & 59.1 & 65.8 & 160 \\
             \hline
             \multirow{4}{*}{\textbf{Ours}} & 0 & 64.5 & 61.8 & 67.2 & 99 \\
             & 2 & 66.3 & 63.3 & 69.4 & 164 \\
             & \cellcolor{blue!7.5}\textbf{4} & \cellcolor{blue!7.5}\textbf{69.8} & \cellcolor{blue!7.5}\textbf{67.0} & \cellcolor{blue!7.5}\textbf{72.7} & \cellcolor{blue!7.5}195 \\
             & 6 & 69.3 & 66.7 & 72.0 & 212 \\
             \hline
        \end{tabular}
    }
    \label{tab:efficiency}
\end{table}

\renewcommand{\arraystretch}{1}
\begin{table}[t]
    \vspace{-2mm}
    \caption{Impact of different MLLMs for Refer-Agent. ``-T'' refers to ``Thinking''.}
    \vspace{-3mm}
    \setlength{\tabcolsep}{3pt}
    \centering
    \resizebox{0.48\textwidth}{!}{
        \begin{tabular}{c|l|cccc}
             \hline
             Methods & MLLM & $\mathcal{J}$\&$\mathcal{F}$ & $\mathcal{J}$ & $\mathcal{F}$ & Time \\
             \hline\hline
             \rowcolor{gray!7.5}
             RGA3~\cite{wang2025object} & Qwen2.5-VL-7B & 53.6 & 51.3 & 56.0 & 32.5 \\\hline
             \multirow{4}{*}{Refer-Agent} & Qwen2.5-VL-7B & 55.5 & 52.4 & 58.6 & 43.1 \\
             & Qwen3-VL-8B-T & 69.1 & 66.2 & 72.1 & 188\\
             & Ovis2.5-9B & 69.8 & 67.0 & 72.7 & 195\\
             \hline
        \end{tabular}
    }
    \label{tab:mllm}
    \vspace{-1em}
\end{table}

\begin{figure*}[!tp]
    \centering
    \captionsetup{type=figure}
    \includegraphics[width=\linewidth]{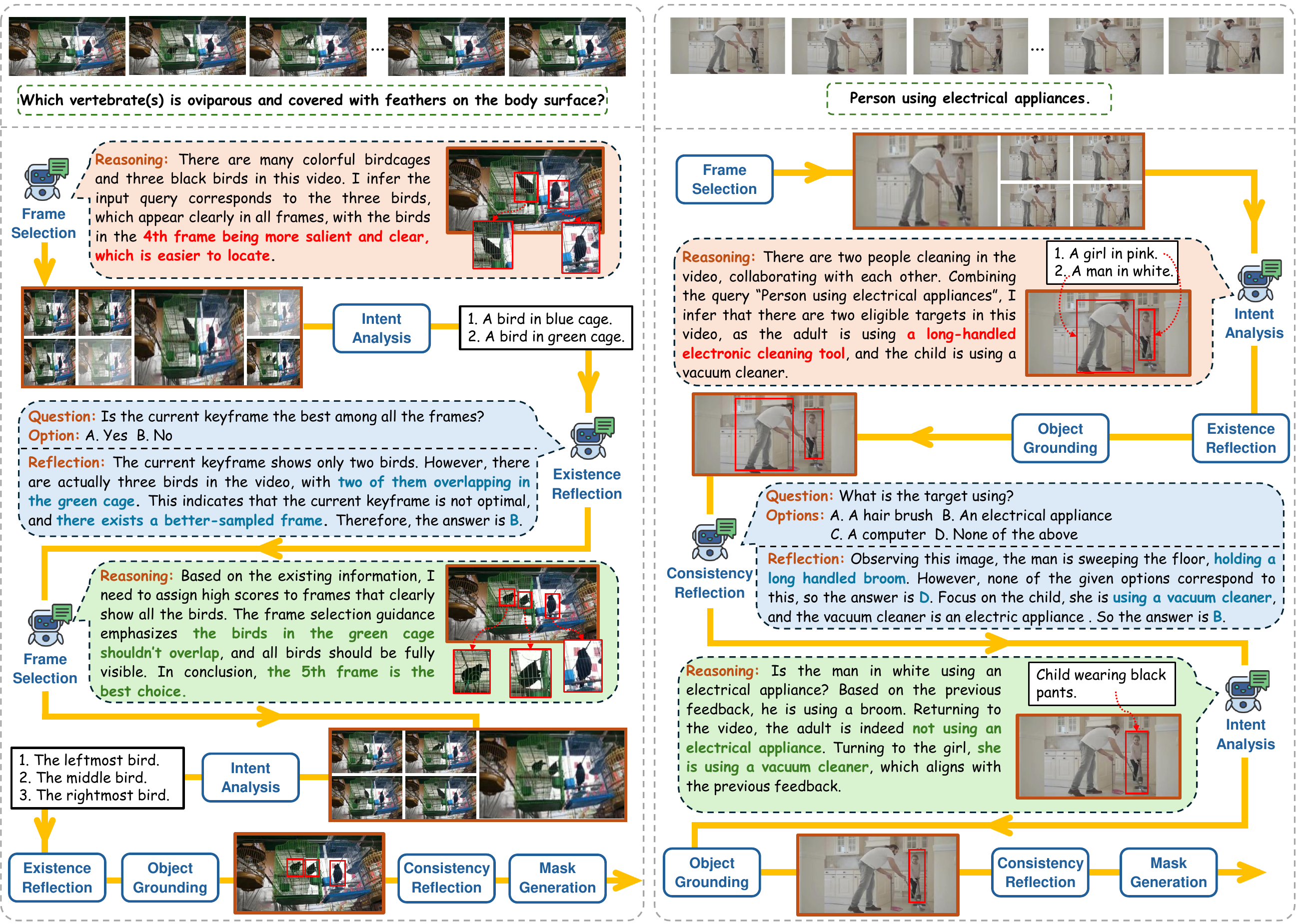}
    \vspace{-1.5em}
    \caption{Visualization of Refer-Agent's alternating reasoning-reflection process. The data flow is highlighted with \textcolor[rgb]{1.0,0.753,0.0}{\textbf{yellow}} arrows. \textbf{Left}: A case of correcting frame selection via Existence Reflection. \textbf{Right}: A case of correcting object identification via Consistency Reflection.}
    \label{fig:reflection}
   \vspace{-0.5em}
\end{figure*}

\subsection{Ablation Studies}
\label{ablation}
Here, we conduct ablation studies to demonstrate the effectiveness of key components in our Refer-Agent. Note that the ablation studies are mainly conducted on ReasonVOS.

\noindent \textbf{Two-stage Chain-of-Reflection.}
We propose a Two-stage Chain-of-Reflection mechanism, comprising \textit{Existence Reflection} (Stage 1) and \textit{Consistency Reflection} (Stage 2), to verify correctness of the step-by-step reasoning procedure. 
As shown in Table~\ref{table:reflection_ablation}, both stages significantly contribute to performance. Ablating either component leads to a substantial performance drop (more than 3\% in $\mathcal{J} \& \mathcal{F}$), while removing both, as expected, yields the poorest results. These results demonstrate the necessity of incorporating a reflection mechanism to rectify the reasoning errors. We also present two cases in Figure~\ref{fig:reflection} to show how the two stages of reflection work (see Sec.~\ref{sec:effectiveness} for details). These results indicate the effectiveness of our Refer-Agent in improving referring video object segmentation through reflection.

\begin{figure*}
    \centering
    \captionsetup{type=figure}  \includegraphics[width=0.97\linewidth]{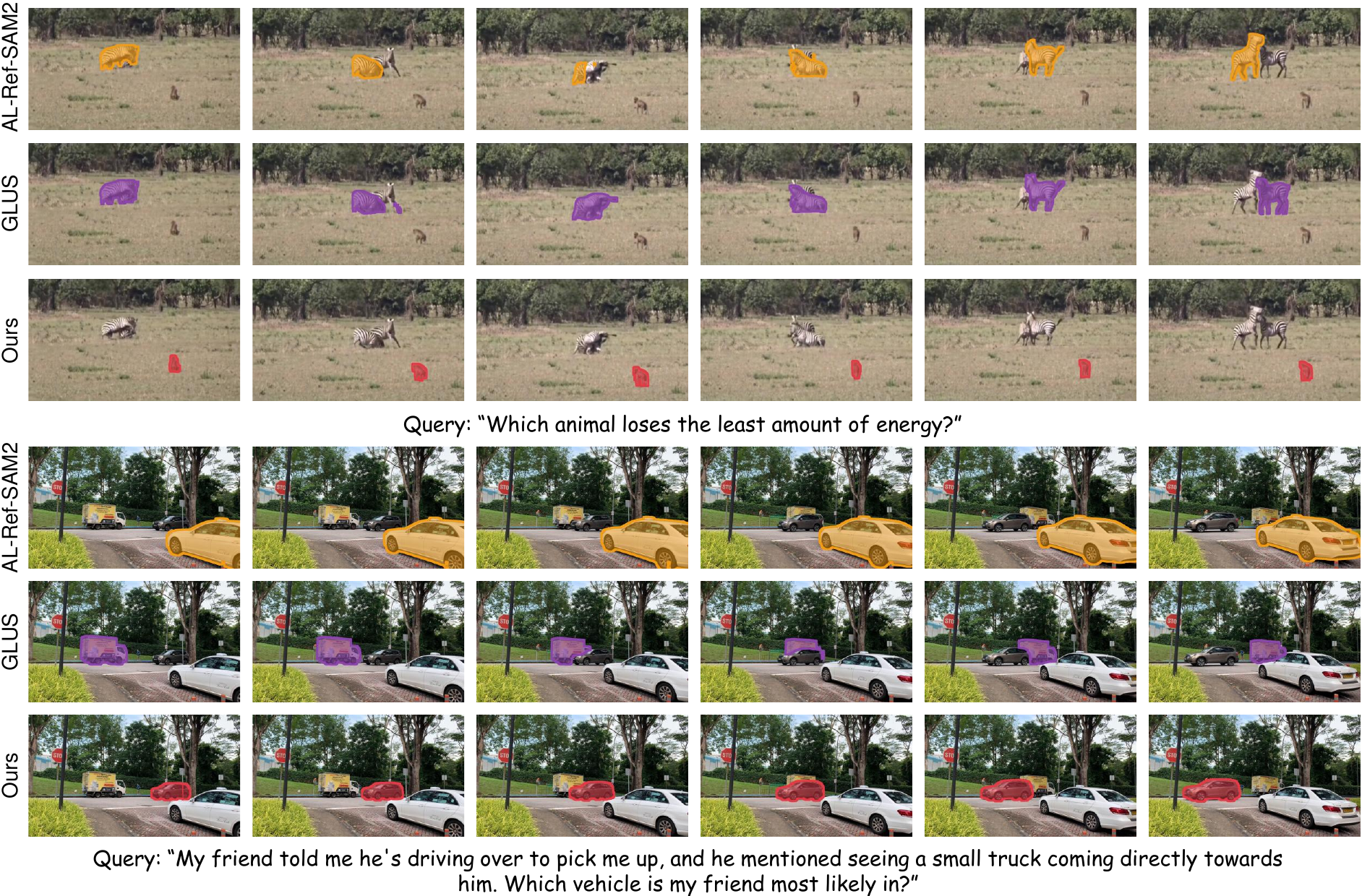}
    \vspace{-1.0em}
    \caption{Qualitative results comparing our Refer-Agent with AL-Ref-SAM2 and GLUS. In the first sample, Refer-Agent correctly segments the small and motionless monkey, while competitors mistakenly segment distant zebras. In the second sample, it accurately segments the black car, whereas AL-Ref-SAM2 and GLUS incorrectly segment the white car and the yellow truck, respectively.}
    \label{fig:visualization}
    \vspace{-0.7em}
\end{figure*} 

\noindent\textbf{Dynamic Focus Layout.}
In Sec.\ref{sec:pipeline}, we propose Dynamic Focus Layout to dynamically adjust frame scales, allowing MLLM to focus on the keyframe while preserving a coherent understanding of temporal dynamics. Here, we evaluate its effectiveness by replacing it with strategies used in previous methods: 1) reasoning with a single keyframe~\cite{huang2024unleashing}; and 2) integrating all sampled frames uniformly~\cite{CoTRVS}. As shown in Table~\ref{tab:input_strategy}, our Dynamic Focus Layout achieves 2.8\% and 2.1\% improvements in $\mathcal{J} \& \mathcal{F}$ over the alternatives, respectively, clearly verifying its effectiveness.

\noindent\textbf{Model Efficiency.}
The efficiency of Refer-Agent mainly depends on the number of reflections (i.e., \texttt{MAX\_TURN}). Here, we explore its impact on the overall performance and efficiency. 
As shown in Table~\ref{tab:efficiency}, Refer-Agent without reflection is efficient but yields the poorest results, achieving only $64.5\%$ in $\mathcal{J}\&\mathcal{F}$. Increasing \texttt{MAX\_TURN} to 4 results in a $5.3\%$ improvement in $\mathcal{J}\&\mathcal{F}$, demonstrating that our model benefits significantly from the reflection process. However, further increasing \texttt{MAX\_TURN} slightly decreases accuracy while increasing the inference cost. Hence, we set \texttt{MAX\_TURN} as 4 to balance performance and efficiency.

We further analyze the impact of our architecture designs on efficiency. In the primary reasoning pipeline, Refer-Agent sequentially employs three agents for frame selection, intent analysis, and object grounding. Intuitively, the overall efficiency could be further improved by merging some of the agents and using a single MLLM to process multiple tasks at once. To study its effects, we implement three variants of Refer-Agent by merging different agents: 1) frame selection and intent analysis; 2) intent analysis and object grounding; and 3) all tasks combined. As shown in Table~\ref{tab:efficiency}, while agent merging brings efficiency gains, it could significantly degrade the overall performance. Under the same \texttt{MAX\_TURN}, our Refer-Agent achieves the best performance while maintaining comparable efficiency. These results confirm the effectiveness of our multi-agent designs in balancing the performance and efficiency.

\noindent\textbf{Impacts of different MLLMs.} We explore the effects of using various MLLMs in our framework. As shown in Table \ref{tab:mllm}, our Refer-Agent is compatible with different MLLMs. A larger MLLM yields superior performance but introduces longer reasoning chains and heavier computation. Using the same MLLM (i.e., Qwen-2.5VL-7B), our Refer-Agent without any fine-tuning still outperforms the SFT-based method RGA3~\cite{wang2025object} with only $\sim$10s extra latency, validating the superiority of our zero-shot framework.

\subsection{Qualitative Analysis}
\label{sec:qualitative}
\textbf{Effectiveness of Reflection.}\label{sec:effectiveness} We present two visualized cases to demonstrate the effectiveness of our Chain-of-Reflection mechanisms. \underline{In the left panel}, the Existence Reflection agent identifies that the keyframe contains only two black birds, while the video actually contains three. Therefore, the agent concludes that the selected keyframe is suboptimal and instructs the Frame Selection agent to refine the keyframe selection. \underline{In the right panel}, the Consistency Reflection agent identifies that the man is not using an electrical appliance, and instructs the Intent Analysis agent to correct the reasoning results. These cases highlight the robust self-correction capability of our Refer-Agent. More details are provided in the Supplementary.

\noindent\textbf{Qualitative Comparison with SOTAs.} 
In Figure~\ref{fig:visualization}, we compare the qualitative results of our Refer-Agent with the zero-shot method AL-Ref-SAM2~\cite{huang2024unleashing} and SFT-based model GLUS~\cite{lin2025glus}. 
As shown, our Refer-Agent achieves the best performance in segmenting objects referenced by the queries.
\underline{In the first sample}, which aims to segment the monkey closest to the camera observing the two fighting zebras. We find that our model can accurately segment the right object, whereas competitors incorrectly segment the zebras in the distance. \underline{In the second sample}, our method can correctly understand this complex reasoning query and infer that the target is the black car heading towards the truck, while AL-Ref-SAM2 and GLUS incorrectly segment the white car and the truck parked by the roadside, respectively. These results demonstrate the effectiveness of our approach for segmenting objects in different scenarios. Please refer to supplementary for more results.

\section{Conclusion}
\label{conclusion}
\vspace{-0.1em}
We introduce Refer-Agent, a novel training-free collaborative multi-agent system for Referring Video Object Segmentation (RVOS), which incorporates a Coarse-to-Fine frame selection approach and a Dynamic Focus Layout strategy to enhance the reasoning segmentation. We also propose a Two-stage Chain-of-Reflection mechanism alternating between reasoning and reflection to mitigate MLLM's hallucinations and enhance the system robustness. Extensive experiments across five public benchmarks validate the effectiveness of our Refer-Agent. 
Furthermore, we demonstrate that our method is flexible and can be seamlessly integrated with existing pre-trained MLLMs.

{
    \small
    \bibliographystyle{ieeenat_fullname}
    \bibliography{main}
}



\end{document}